\documentclass{ceurart}

\usepackage{listings}
\usepackage{listings}
\usepackage{xcolor}
\newcommand{\mintinline}[2]{\lstinline[language=#1]{#2}}

\usepackage{amsmath}
\DeclareMathOperator{\prob}{{P}}

\begin{document}
\copyrightyear{2024}
\copyrightclause{Copyright for this paper by its authors. Use permitted under Creative Commons License Attribution 4.0 International (CC BY 4.0).}
\conference{AIMMES 2024 Workshop on AI bias: Measurements, Mitigation, Explanation Strategies $\vert$ co-located with EU Fairness Cluster Conference 2024, Amsterdam, Netherlands}

\title{Towards Standardizing AI Bias Exploration}
\author{Emmanouil Krasanakis}[%
orcid=0000-0002-3947-222X,
email=maniospas@iti.gr,
]
\cormark[1]
\author{Symeon Papadopoulos}[%
orcid=0000-0002-5441-7341,
email=papadop@iti.gr,
]
\address{Centre for Research \& Technology Hellas, 6th km Charilaou-Thermi, Thessaloniki, Greece, 57001}
\cortext[1]{Corresponding author.}

\begin{abstract}
Creating fair AI systems is a complex problem that involves the assessment of context-dependent bias concerns. Existing research and programming libraries express specific concerns as measures of bias that they aim to constrain or mitigate. In practice, one should explore a wide variety of (sometimes incompatible) measures before deciding which ones warrant corrective action, but their narrow scope means that most new situations can only be examined after devising new measures. In this work, we present a mathematical framework that distils literature measures of bias into building blocks, hereby facilitating new combinations to cover a wide range of fairness concerns, such as classification or recommendation differences across multiple multi-value sensitive attributes (e.g., many genders and races, and their intersections). We show how this framework generalizes existing concepts and present frequently used blocks. We provide an open-source implementation of our framework as a Python library, called \textit{FairBench}, that facilitates systematic and extensible exploration of potential bias concerns.
\end{abstract}


\maketitle

\section{Introduction}
Artificial Intelligence (AI) systems see widespread adoption across many applications that affect people's lives. Since they tend to pick up and exacerbate real-world biases or discrimination, as well as spurious correlations between predicted values and sensitive attributes (e.g., gender, race, age, financial status), making them fair is a subject of intensive research and regulatory efforts. These include quantification of bias concerns through appropriate measures so that unfair behavior can be detected and corrected. To this end, several measures of bias have been proposed in research papers and implemented as reusable components within programming libraries or toolkits (Section~\ref{background}). Recognizing that bias and, more generally, fairness depends on the social context and the particular settings in which AI systems are deployed, each created measure is limited to assessing a different type of concern. In the end, research and development focuses on mitigating or constraining measures when those reveal fairness issues, but not on how a systematic exploration of many measures could be carried out to perform fairness audits.

Therefore, and despite the obvious value of presenting reusable algorithmic solutions to specific problems, there is a need for methods that critically examine real-world systems across a wide range of concerns and not just a few of them. The main barrier in pursuing such methods is that measures of bias are designed in a monolithic manner and rarely consider how they could be generalized or ported to different settings. For example, differential fairness \cite{foulds2020intersectional} was only recently proposed as a means of generalizing disparate impact assessment to intersectional fairness, which recognizes the cumulative effect of sensitive attributes (e.g., multiple genders and races), despite disparate impact measures being around for decades \cite{biddle2017adverse}. 

In this work, we assist systematic exploration of fairness concerns by decomposing measures of bias into simple building blocks. These can be recombined to create new measures covering a wide range of contexts and concerns. For example, a block that aggregates classification bias in multidimensional settings (e.g., with multiple races and genders) can be combined with a block that another measure uses to assess recommendation bias between only two groups of people (e.g., only whites vs non-whites) to create a new measure that assesses multidimensional recommendation bias. 
We implement the proposed framework in a Python library, called \textit{FairBench}, that standardizes how measures of bias are defined by combining interoperable blocks of each kind. The library's functional interface sets up a fixed representation of existing and new measures of bias, and can create bias reports while tracing prospective fairness issues to the raw quantities computed over predictive outcomes. Our contribution is threefold:
\begin{itemize}
    \item[a)] We present a mathematical framework that systematically combines building blocks to construct many existing and new measures of bias.
    \item[b)] We express several building blocks of literature measures of bias within this framework.
    \item[c)] We introduce the FairBench Python library that implements the above blocks and combination mechanisms to compute measures of bias in a wide range of computing environments.
\end{itemize}

This paper is structured as follows. After this section's introduction, Section~\ref{background} presents theoretical background and related work. Section~\ref{mathematical framework} describes our proposed mathematical framework. Section~\ref{building blocks} extracts several bias building blocks from the AI fairness literature. Section~\ref{fairbench} introduces the programming interface provided by our FairBench Python library to explore bias. 
Finally, Section~\ref{conclusions} summarizes our work and points to future directions.

\section{Background and Related Work}\label{background}
\subsection{Fair AI}
The problem of creating fair AI systems is the subject has attracted attention as part of the broader theme of Trustworthy AI in worldwide regulation and ethical guidelines, like the EU's Assessment List for Trustworthy AI \cite{ala2020assessment}, and the NIST's AI Risk Management Framework \cite{ai2023artificial}. Evaluating system fairness is a complex problem that depends on the context being examined and the systems being created. A part of answering this problem consists of mathematical or algorithmic exploration that enables automated system assessment and oversight with practices like TrustAIOps \cite{li2023trustworthy}, which monitor the evolution of system bias in the deployment context.

Mathematical definitions of fairness are often categorized into the following types \cite{ntoutsi2020bias,barocas2023fairness,mitchell2021algorithmic,mehrabi2021survey}:
\begin{itemize}
    \item[i)] \textit{Group fairness} focuses on equal treatment between population groups or subgroups. This is the subject of most research and covered extensively in the next subsection.
    \item[ii)] \textit{Individual fairness} \cite{dwork2012fairness} focuses on the fair treatment of individuals, for example by obtaining similar predictions for those with similar features. This can be modelled as group fairness too, by considering every person to belong to their own group.
    \item[iii)] \textit{Counterfactual fairness} \cite{kusner2017counterfactual,carey2022causal} learns causal models of predictive mechanisms that account for discrimination and then makes predictions in a would-be fair reality. Although originally coined as a variation of individual fairness, recent understanding \cite{rosenblatt2023counterfactual} also suggests that counterfactual fairness is a kind of group fairness.
\end{itemize}

Making (e.g., training) AI to be fair typically involves measures of bias that quantify numerical deviation from exact definitions of fairness. Measures are either minimized or subjected to constraints \cite{xinying2023guide}, but identifying which are important is not only a matter of applying scientific principles. In particular, it is mathematically impossible to simultaneously satisfy all conceivable definitions of fairness \cite{kleinberg2016inherent,miconi2017impossibility}, which means that systems should only address the fairness concerns that matter to humans (e.g., stakeholders or policymakers) in the particular situation. There are also trade-offs between satisfying fairness and maintaining predictive performance. 
In this work we propose that many types of bias should be monitored simultaneously to reveal concerning trends that require further context-dependent human evaluation. 

In practice, AI systems employ fairness libraries or toolkits to compute popular measures of bias and either constrain predictive tasks to achieve the accepted measure values or carry out trade-offs of the latter with predictive performance. 
Some popular software projects for fair AI are AIF360 \cite{bellamy2019ai}, FairLearn \cite{bird2020fairlearn}, What if \cite{wexler2019if}, and Aequitas \cite{saleiro2018aequitas}. Each of these focuses on supporting different computational backends and data structures. AIF360 and FairLearn are programming libraries and provide bias mitigation algorithms. What If and Aequitas focus on assessment of fairness, and specialize in assessing counterfactual and group fairness respectively. All libraries and toolkits implement ad-hoc measures of bias that capture fairness concerns well-studied in the literature, where the latter typically account for very restricted types of evaluation that do not model complex concerns.


\subsection{Measures of bias}\label{background measures}
We now present common measures of bias, starting from ones that quantify group fairness concerns for classifiers \cite{castelnovo2021zoo}. We organize people with the same sensitive attribute values into population groups. All presented measures assume values in the range $[0,1]$ and, when necessary, we show the complements of measures of fairness with respect to $1$ to let them assess bias instead, i.e., a value of $0$ indicates perfect fairness. For example, below we use $1-\text{prule}$ as a measure of bias, instead of $\text{prule}$, which is a measure of fairness.

\paragraph{Measures of classification bias.} The measures of $1-\text{prule}$ \cite{biddle2017adverse} and Calders-Verwer disparity $\text{cv}$ \cite{calders2010three} quantify disparate impact concerns, i.e., prediction rate inequality between two groups of samples $\mathcal{S}$ and $\mathcal{S}'=\mathcal{S}_{all}\setminus\mathcal{S}$, where the latter complements the former within the population $\mathcal{S}_{all}$. This models one binary sensitive attribute indicating membership to the group. Mathematically:
$$\text{prule}=\min\Big\{\tfrac{\prob(c(x)=1|x\in \mathcal{S})}{\prob(c(x)=1|x\in \mathcal{S}')},\tfrac{\prob(c(x)=1|x\in \mathcal{S}')}{\prob(c(x)=1|x\in \mathcal{S})}\Big\}\,,\, \text{cv}=|\prob(c(x)=1|x\in \mathcal{S})-\prob(c(x)=1|x\in \mathcal{S}')|$$
where $\prob(\cdot|\cdot)$ denotes conditional probability and $c(x)\in\{0,1\}$ is the binary outcome of classifying data sample $x$. Disparate impact is eliminated when $1-\text{prule}=\text{cv}=0$. 
Another concern is disparate mistreatment \cite{zafar2017fairness}, which captures misclassification differences between groups per:
\begin{equation*}\label{differences}
    |\Delta m|=|m(\mathcal{S})-m(\mathcal{S}')|
\end{equation*}
where $m(\cdot)$ denotes a misclassification measure over groups of samples, such as their false positive rate (fpr) or their false negative rate (fnr). The same formula can express $\text{cv}$ or accommodate error measures for other predictive tasks, and therefore constitutes a building block of generalized fairness evaluation frameworks \cite{roy2023multi} (also see Subsection~\ref{comparisons}). An example of difference-based bias for probabilistic measures of predictive performance is equalized opportunity difference \cite{hardt2016equality,donini2018empirical}:
$$|\Delta eo|=|\prob(c(x)=1|x\in \mathcal{S}, Y(x)=y)-\prob(c(x)=1|x\in \mathcal{S}',Y(x)=y)|$$
where $y\in\{0,1\}$ and $Y(x)$ is the true test/validation data label corresponding to sample $x$.

\paragraph{Measures of scoring and recommendation bias.}
The same underlying principles can be ported to other predictive tasks \cite{li2023trustworthy,xinying2023guide} such as recommendation and scoring. In recommendation, the base measure $m(\cdot)$ in $|\Delta m|$ may be replaced by some form of exposure of items to group members or the fraction of top-$k$ recommendations that are members of the group. In scoring tasks, $m(\cdot)$ may represent the fraction of score mass concentrated on groups, which can be compared to a desired fraction or between groups to satisfy concerns similar to disparate impact \cite{tsioutsiouliklis2021fairness}. Aggregate statistics, like the area under the roc (auc) or score means, may also be compared between groups \cite{calders2013controlling}. A definition that we use later for receiver operating characteristic (roc) curves of groups $\mathcal{S}_i$ accounts for all pairs of false positive rate (fpr) and true positive rate (tpr) values obtained for the groups over different thresholds $\theta$ of whether scores are interpreted as positive predictions or not (we use the maplet arrow to express rocs as maps from fpr to tpr): 
\begin{equation}\label{roc}
\text{roc}(\mathcal{S}_i)=\{\text{fpr}(\mathcal{S}_i,\theta)\mapsto \text{tpr}(\mathcal{S}_i,\theta):\theta\in(-\infty, \infty)\}
\end{equation}

More fine-grained approaches summarize the differences between curves or distributions (a mathematical expression for this appears in Subsection~\ref{comparisons}). For example, viable measures of bias are the absolute betweenness area between roc curves (abroca) \cite{gardner2019evaluating} and the Kullback-Leibler divergence between distributions of system outcomes for each group \cite{ghosh2021characterizing}. 

\subsection{Frameworks to assess multidimensional bias}\label{multidiscrimination frameworks}
Fairness concerns may span several sensitive groups and their intersections \cite{kearns2018preventing}. This setting is known as multidimensional fairness. For example, there may be multiple protected demographics (e.g., genders, races, and their intersections). 
Two frameworks for expressing several measures of bias in the multidimensional setting are a) what we later call \textit{groups vs all} comparison \cite{roy2023multi} and b) \textit{worst-case bias} \cite{ghosh2021characterizing}. In this work, we generalize these frameworks under a more expressive one. Our analysis also accounts for c) individual fairness by treating it under a multidimensional framework where each individual is a separate group of one element.

\paragraph{Groups vs all.} This sets up the following generic framework for measures of bias:
\begin{equation}\label{roy}
    F_{bias}(\mathbb{S})=\odot_{\mathcal{S}_i\in\mathbb{S}}\, F_{base}(\mathcal{S}_i)
\end{equation}
where $\mathbb{S}$ are all groups, $F_{base}(\mathcal{S}_i)$ is a measure of bias for one group $\mathcal{S}_i\in\mathbb{S}$ (this corresponds to treating that group as having only one binary sensitive attribute), and $\odot$ is a reduction mechanism, such as the minimum or maximum. Groups may be overlapping when examining each sensitive attribute value independently without considering their intersections \cite{kearns2019empirical,ma2021subgroup,martinez2021blind,shui2022learning}. It has been proposed that group intersections may also replace the groups within this analysis, therefore creating \textit{subgroups} in their place \cite{foulds2020intersectional,kearns2018preventing,ghosh2021characterizing}. 

In Equation~\ref{roy}, the base measure of bias compares one group against the total population. Example measures that employ this scheme are statistical parity subgroup fairness (spsf) and false positive subgroup fairness (fpsf) \cite{kearns2018preventing}. For these, each subgroup is compared to the total population to create multidimensional variations of $\Delta eo$ and $\Delta \text{fpr}$ respectively. Both employ a reduction mechanism that weighs comparisons by group size.

\paragraph{Worst-case bias.} This framework starts from pairwise group or subgroup comparisons and keeps the worst case. We compute worst-case bias (wcb) for any probabilistic measure $F(\mathcal{S}_i)$, such as of accurate or erroneous predictions over groups or subgroups $\mathcal{S}_i\in\mathbb{S}$, per:
\begin{equation}\label{wcf}
    \text{wcb}(\mathbb{S}) = 1-\min_{(\mathcal{S}_i,\mathcal{S}_j)\in\mathbb{S}^2} \tfrac{F(\mathcal{S_i})}{F(\mathcal{S_j})}
\end{equation}
For example, differential fairness \cite{foulds2020intersectional} states that prule should reside in the range $[e^{-\epsilon},e^\epsilon]$ for some $\epsilon\geq 0$ when it compares \textit{all} subgroups pairwise. A viable measure of bias for this definition, which we call differential bias (db), is given for subgroups $\mathbb{S}$ when $F(\mathcal{S}_i)=\prob(c(x)=1|x\in\mathcal{S}_i)$, for which differential fairness is equivalent to satisfying $\text{wcb}(\mathbb{S})\leq 1-e^{-\epsilon}$.

\paragraph{Individual fairness} Multidimensional discrimination can also model individual fairness \cite{dwork2012fairness,kim2019preference} by setting each individual as a separate group. In this case, and given that individual fairness is formally defined to satisfy $d_y(c(x_i),c(x_j))\leq d_x(x_i,x_j)$
across all pairs of individuals $(x_i,x_j)$, where $c$ is a predictive mechanism and $d_y,d_x$ are some distance metrics, a viable measure of individual bias (ib) that satisfies individual fairness when $\text{ib}\leq 1$ is:
\begin{equation}
    \text{ib}=\max_{x_i,x_j}\tfrac{d_y(c(x_i),c(x_j))}{d_x(x_i,x_j)}
\end{equation}

\section{A General Bias Measurement Framework}\label{mathematical framework}
In this section we present a framework for defining measures of bias from fundamental building blocks. This lets us decompose existing measures into blocks, introduce variations, and create novel combinations. We recognize four types of blocks, which correspond to successive computational steps: a) selecting which pairs of population (sub)groups to compare, b) base measures that assess some system property on each group, c) comparisons between group assessments, and d) reductions that convert multiple pairwise comparisons to one value. 

Mathematically, we annotate base measures as $F(\mathcal{S}_i)$ and compute them over groups of data samples $\mathcal{S}_i\subseteq \mathcal{S}_{all}$ of some population $\mathcal{S}_{all}$. We consider any information necessary for this computation (e.g., predicted and ground truth labels) directly retrievable from the respective samples. We also annotate the set of all these groups as $\mathbb{S}=\{\mathcal{S}_i:i=0,1,\dots\}$. The base measures could be error rates of predictions or unsupervised statistics, like positive rates. Then, we consider a set of subgroup pairs to compare $C(\mathbb{S},\mathcal{S}_{all})\in \big(2^{\mathcal{S}_{all}}\big)^2$, where $2^{\mathcal{S}_{all}}$ denotes the powerset. The created set of subgroup pairs comprises any necessary detected comparisons between groups or subgroups within $\mathcal{S}_{all}$, even those that do not directly reside in $\mathbb{S}$ but may be derived from the latter, such as subgroups. We will make pairwise comparisons $f_{ij}=F(\mathcal{S}_i)\oslash F(\mathcal{S}_j)$ between (sub)groups $(\mathcal{S}_i,\mathcal{S}_j)\in C(\mathbb{S},\mathcal{S}_{all})$. Finally, the reduction mechanism will be expressed as $\odot_{(S_i,S_j)}f_{ij}$ across all pairs $(S_i,S_j)$ and outcomes of comparing them $f_{ij}$.

Putting the above in one formula yields our framework for expressing measures of bias:
\begin{equation}\label{framework}
    F_{bias}=\odot_{(\mathcal{S}_i,\mathcal{S}_j)\in C(\mathbb{S},\mathcal{S}_{all})}\, \big(F(\mathcal{S}_i)\oslash F(\mathcal{S}_j)\big)
\end{equation}
This is a direct generalization of Equation~\ref{roy} in that it supports more varied strategies for expressing group comparisons. At the same time, it systematizes the comparison mechanisms between groups of people through the operation $\oslash$. 
Our framework is also a generalization of Equation~\ref{wcf} in that, in addition to the base measure (which we allow to also be non-probabilistic, if so desired), it provides flexibility in the reduction and comparison mechanisms beyond the choices of minimum and fractional comparison that characterize the worst case.

Each tuple of choices $(F, C,\oslash,\odot)$ creates a different measure of bias. In the next section, we delve into how each building block type among the members of this tuple could vary between contexts that comprise different fairness concerns and predictive tasks. For every building block, a systematic exploration would consider all possible combinations with others of different kinds so that eventually we not only reconstruct the original measures of bias, but also create variations that address different settings.  Table~\ref{tab:examples} exemplifies how the measures discussed in Subsection~\ref{background measures} arise from specific choices. Creating a full taxonomy of existing measures under our framework is not this work's objective (we only aim to demonstrate its wide expressive breadth) and is left to future work.

\begin{table}[htpb]
    \centering
    \setlength{\tabcolsep}{3pt}
    \begin{tabular}{l r c c c c}
        \textbf{Measure} & ~ & $F(\mathcal{S_i})$ & $C(\mathbb{S},\mathcal{S}_{all})$ & $f_i\oslash f_j$ & $\odot$\\
        \hline
        ~&\multicolumn{5}{c}{Example measures}\\
        \hline
         $|\Delta \text{fpr}|$ & \cite{zafar2017fairness} & $\prob(c(x)=1|x\in\mathcal{S}_i,Y(x)=0)$ & $\{\mathcal{S},\mathcal{S}_{all}\setminus \mathcal{S}\}^2$ for $\mathbb{S}=\{\mathcal{S}\}$ & $f_i-f_j$ & $\max$ \\
         $|\Delta \text{fnr}|$ & \cite{zafar2017fairness} & $\prob(c(x)=0|x\in\mathcal{S}_i,Y(x)=1)$ & $\{\mathcal{S},\mathcal{S}_{all}\setminus \mathcal{S}\}^2$ for $\mathbb{S}=\{\mathcal{S}\}$ & $f_i-f_j$ & $\max$ \\
         $|\Delta eo|$ & \cite{hardt2016equality} & $\prob(c(x)=1|x\in\mathcal{S}_i)$ & $\{\mathcal{S},\mathcal{S}_{all}\setminus \mathcal{S}\}^2$ for $\mathbb{S}=\{\mathcal{S}\}$ & $f_i-f_j$ & $\max$\\
         $\text{spsf}$ & \cite{kearns2018preventing} & $\prob(c(x)=1|x\in\mathcal{S}_i)$ & $\mathbb{S}\times \{\mathcal{S}_{all}\}\cup \{\mathcal{S}_{all}\}\times\mathbb{S}$ & $|f_i-f_j|$ & wmean\\
         $\text{fpsf}$ & \cite{kearns2018preventing} & $\prob(c(x)=1|x\in\mathcal{S}_i,Y(x)=0)$ & $\mathbb{S}\times \{\mathcal{S}_{all}\}\cup \{\mathcal{S}_{all}\}\times\mathbb{S}$ & $|f_i-f_j|$ & wmean\\
         $\text{cv}$ & \cite{calders2010three} & $\prob(c(x)=1|x\in\mathcal{S}_i)$ & $\{\mathcal{S},\mathcal{S}_{all}\setminus \mathcal{S}\}^2$ & ${f_i}-{f_j}$ & $\max$\\
         $1-\text{prule}$ & \cite{biddle2017adverse} & $\prob(c(x)=1|x\in\mathcal{S}_i)$ & $\{\mathcal{S},\mathcal{S}_{all}\setminus \mathcal{S}\}^2$ & $1-{f_i}/{f_j}$ & $\max$\\
         $\text{db}$ & \cite{foulds2020intersectional} & $\prob(c(x)=1|x\in\mathcal{S}_i)$ & $\mathbb{S}^2$ & $1-{f_i}/{f_j}$ &  $\max$\\
         \hline
         ~&\multicolumn{5}{c}{Multidimentional bias frameworks}\\
         \hline
         $F_{bias}$ & \cite{roy2023multi} & any$^*$ & $\mathbb{S}\times \{\mathcal{S}_{all}\}\cup \{\mathcal{S}_{all}\}\times\mathbb{S}$ & any &  any\\
         $\text{wcb}$ & \cite{ghosh2021characterizing} & any & $\mathbb{S}^2$ & $1-{f_i}/{f_j}$ &  $\max$\\
         $\text{ib}$ & \cite{dwork2012fairness} & $c(x)$ & $\{\{x\}:x\in\mathcal{S}_{all}\}$ & $d_y(f_i,f_j)/d_x(\mathcal{S}_i,\mathcal{S}_j)$ & $\max$
    \end{tabular}
    \caption{Expressing the measures of classification bias of Subsection~\ref{background measures} and multidimensional bias frameworks of Subsection~\ref{multidiscrimination frameworks} under our bias measure definition framework. Interpretation of reductions can be found in Subsection~\ref{reductions}. Frameworks allow for any blocks of certain kinds. $^*$$F_{bias}$ does not explicitly acknowledge base measure building blocks.}
    \label{tab:examples}
\end{table}

\section{Bias building blocks}\label{building blocks}
Here we present building blocks that occur from decomposing popular measures of bias of Sections~\ref{background measures} and~\ref{multidiscrimination frameworks}. Combinations under our framework can create new measures.

\subsection{Selecting groups or subgroups to compare}\label{selecting}
The group selection mechanisms we study include a) base selection, b) accounting for individual fairness \cite{dwork2012fairness}, and c) accounting for intersectionality. 
More blocks can be created in the future. 

\paragraph{Base selection.}
Several fairness measures are defined by comparing each population group or subgroup with the total population $\mathcal{S}_{all}$. Given that the pairwise comparison mechanism is not symmetric, i.e., it may hold that $f_i\oslash f_j\neq f_j\oslash f_i$, we account for both $(\mathcal{S}_i,\mathcal{S}_{all})$ and $(\mathcal{S}_{all},\mathcal{S}_i)$ for subgroups $\mathcal{S}_i\in\mathbb{S}$. Mathematically, we define a \textit{group vs any sample} selection per:
\begin{equation}\label{vsAny}
    \text{vsany}(\mathbb{S},\mathcal{S}_{all}) = \mathbb{S}\times\{\mathcal{S}_{all}\}\cup\{\mathcal{S}_{all}\}\times \mathbb{S}
\end{equation}
Alternatively, one may consider all group pairs (we let reduction remove any self-comparisons):
\begin{equation}\label{allvsall}
    \text{pairs}(\mathbb{S},\mathcal{S}_{all}) = \mathbb{S}^2
\end{equation}

For one-dimensional fairness, i.e., only one sensitive attribute with only two potential values, such as indicating which samples belong to a protected group of people and which do not, each group $\mathcal{S}$ is typically compared to its \textit{complement} within the whole population $\mathcal{S}_{all}\setminus\mathcal{S}_i$. In this case, $\mathbb{S}=\{\mathcal{S}\}$ and we write the pairwise selection between it and its complement per
$
    \text{compl}(\mathbb{S},\mathcal{S}_{all}) = \{\mathcal{S},\mathcal{S}_{all}\setminus \mathcal{S}\}^2
$.
One generalization to multidimensional settings would be to consider $\mathcal{S}_{all}\setminus \mathcal{S}$ as a group and recreate Equation~\ref{allvsall}. However, an equally valid generalization it to compare groups to their complements only:
\begin{equation}
    \text{compl}(\mathbb{S},\mathcal{S}_{all}) = \bigcup_{\mathcal{S}_i\in\mathbb{S}}\{\mathcal{S}_i,\mathcal{S}_{all}\setminus \mathcal{S}_i\}^2
\end{equation}



\paragraph{Accounting for individual fairness.}
Individual fairness disregards the notion of population groups or subgroups and instead compares all individuals pairwise. This is an instance of the $\text{pairs}$ mechanism, where each individual is assigned to their own group. Mathematically, this would occur if we specified subgroups $\mathbb{S}=\{\{x\}:x\in\mathcal{S}_{all}\}$. We hereby extend all comparison mechanisms $C$ to follow this schema when an empty set of groups $\emptyset$ is provided\footnote{The empty set indicates no knowledge of a group, in which case our fallback becomes individual fairness.} per:
\begin{equation}
   C(\emptyset,\mathcal{S}_{all})=C(\{\{x\}:x\in\mathcal{S}_{all}\},\mathcal{S}_{all}) 
\end{equation}

\paragraph{Accounting for intersectionality.} Measures of bias like $\text{db}$, $\text{spsf}$, and $\text{spsf}$ account for subgroups that form intersections of groups too. To model this scenario, we define intersectional comparisons between groups while ignoring empty intersections (this conveniently ignores intersections between mutually exclusive sensitive attribute values) per:
\begin{equation}\label{cintersect}
    C_{intersect}(\mathbb{S},\mathcal{S}_{all})=C(\{\mathcal{S}=\mathcal{S_{i_1}}\cap \mathcal{S_{i_2}}\cap \dots :\mathcal{S_{i_k}}\in\mathbb{S},\mathcal{S}\neq \emptyset\}, \mathcal{S}_{all})
\end{equation}

\subsection{Base measures}\label{base measures}
Our framework accepts any measure of predictive performance. In this subsection, we demonstrate two concepts: a) computing aggregate assessments to feed as-are in comparison mechanisms, and b) tracking underlying metadata that could be used by other building blocks without altering Equation~\ref{framework}. Base measures can also be ported from fairness definitions that do not specify measures of bias, such as equal group benefit \cite{gb}. Our framework can supply the rest of building blocks to try all available options at defining bias, such as all comparison mechanisms.

\paragraph{Aggregate assessment.}  Here, we list base measures of aggregate system assessment often used to define bias. Probabilistic definitions for them can be found in domain literature. When working with classifiers, popular base measures are positive rates (pr), fpr, and fnr. Other misclassification measures are also used in expressions of disparate mistreatment \cite{zafar2017fairness}. In multiclass settings, each class can be treated as a different group of data samples with its own target labels. Bias has been assessed for the top-$k$ individual/item recommendations by comparing the base measures of hit rate and skew \cite{zehlike2021fairness,pitoura2022fairness} across groups when checking for proportional representation. Recommendation correctness measures may be similarly accommodated, such as the click-through rate \cite{pitoura2022fairness}. Recommendation is assessed across several values of $k$, which leads to measures that keep track of metadata (see below).

\paragraph{Keeping track of metadata.} An emerging concern is that measures of bias affirm the presence of certain biases but not their absence. To verify the latter, one needs to look at nuances of underlying distributions \cite{ghosh2021characterizing}. For example, it has been argued that statistical tests should replace the $\text{prule}$ to show non-discrimination in certain legal settings \cite{watkins2022four}. Thus, our framework retains building block metadata to be used in subsequent computations. To understand this, think of the abroca measure \cite{gardner2019evaluating} that compares the area between the roc curves of Equation~\ref{roc} instead of comparing aucs through differences or fractions. 

We model curve comparisons by tracking computational metadata and retrieving them through an appropriate predicate $\text{curve}(\cdot)$ defined alongside base measures. For example, $\text{auc}$'s definition as a building block should include the statement:
$$\text{curve}(\text{auc}(\mathcal{S}_i))=\text{roc}(\mathcal{S}_i)$$
We do not directly return the metadata (e.g., the roc curves) in order to compare aggregate assessments with as many mechanisms as possible. For instance, also computing $|\text{auc}(\mathcal{S}_i)-\text{auc}(\mathcal{S}_j)|$ and checking differences with abroca may create high-level insights (e.g., if these two quantities are equal between two groups, the corresponding roc curves do not intersect). 

We now present a second base measure used in recommendation systems to assess top predictions and contains a curve, namely average representation (ar) within the top predictions:

$$\text{ar}_K=\tfrac{1}{K}\sum_{k=1}^K\prob(x\in\mathcal{S}_i|x\in \text{top}_k)$$ 
To keep working with integratable curves with continuous horizontal axes, we use Dirac's delta function $\delta(k)=0$ for $k\neq 0$ and $\int_{-\infty}^\infty \delta(k)\text{d}k=1$ to define the injection:
$$\text{curve}(\text{ar}_K)=\big\{k\mapsto\prob(x\in\mathcal{S}_i|x\in \text{top}_k)\delta(0)\text{ if }k\in\{1,2,\dots,K\},0\text{ otherwise}:k\in(-\infty,\infty)\big\}$$

\subsection{Comparison mechanisms}\label{comparisons}
We examine three types of comparison to serve as building blocks in our framework: a) no comparison, b) numeric errors (differences and fractions), and c) curve comparisons. The last type encompasses mechanisms that compute weighed curve differences. Other mechanisms can be readily created, and we demonstrate thresholded variations. 

\paragraph{No comparison.} This is a valid option when aiming to remove confounding bias, i.e., bias that directly leads to erroneous system predictions. For example, the worst measure assessment across groups or subgroups is maximized if all groups exhibit high predictive performance \cite{sarridis2023flac}. In this case, our framework just assesses AI performance for all groups. Mathematically:
$$f_i\oslash_{none} f_j=f_i$$

\paragraph{Numeric errors.} These compute some deviation between $f_i$ and $f_j$, such as absolute error (abs), relative error (rel), and their signed values (sabs and srel respectively):
\begin{align*}
f_i\oslash_{abs} f_j=|f_i-f_j|\,,\,f_i\oslash_{rel} f_j=|1-{f_i}/{f_j}|\,,\,f_i\oslash_{sabs} f_j=f_i-f_j\,,\,f_i\oslash_{srel} f_j=1-{f_i}/{f_j}
\end{align*}

\paragraph{Curve comparisons.} Let us consider that the curve predicate can extract from base measure assessment curves $crv_i=\text{curve}(f_i)$, $crv_j=\text{curve}(f_j)$ with a known common domain $\mathcal{D}$. We can compare these given a weighting mechanism $I(k)$ for the domain and comparison $\oslash_{crv}$ as:
$$f_i\oslash f_j=\tfrac{1}{\int_{\mathcal{D}} I(k)\text{d}\theta}\int_{\mathcal{D}} I(k)\cdot\big(crv_i(k)\oslash_{crv} crv_j(k)\big)\,\text{d}k$$
Essentially, we are building curve comparisons from the simpler components $(I,\oslash_{crv})$. As an example, we reconstruct the abroca pairwise group comparison \cite{gardner2019evaluating} based on the auc base measure of Subsection~\ref{base measures} given the curve comparison defined by the pair of operations $\oslash_{abroca}=(I_{const}(k)=1,\oslash_{abs})$.
Measures that retain curves of top-$k$ recommendations may also consider NDCG-like weighting $I_{DCG}(k)=\big(\tfrac{1}{\log(1+k)}\text{ if }k>0,0\text{ otherwise}\big)$ \cite{zehlike2021fairness,pitoura2022fairness}. 

\paragraph{Thresholded variations.} Oftentimes, small deviations from perfect fairness are accepted. Settling for thresholds under which measures of bias output $0$ can take place either at the end of the assessment, or during intermediate steps. If the last option is chosen, comparisons $f_i\oslash f_j$ need to be replaced with the following variations of maximum acceptable threshold $\epsilon\geq 0$:
\begin{equation*}
    f_i\oslash_{\epsilon} f_j=\max\{0,f_i\oslash f_j-\epsilon\}
\end{equation*}

\subsection{Reduction mechanisms}\label{reductions}
The final step of our framework consists of reducing to one quantity all pairwise comparisons $f_{ij}$ between groups or subgroups $(\mathcal{S}_i,\mathcal{S}_j)\in C(\mathbb{S},\mathcal{S}_{all})$. Common reductions are the maximum, minimum, or arithmetic mean of all values. Especially the maximum is a cornerstone of the worst-case bias framework, although it does not differentiate between systems that have the same worst case. Alternatively, reduction could adopt some formula that becomes the weighted mean in the group vs all case \cite{kearns2018preventing}. Here, we demonstrate one feasible option, which obtains weights $1-|\prob(x\in S_i)-\prob(x\in S_{all})=1-|\prob(x\in S_i)-1|=\prob(x\in S_i)$ under $\text{vsany}$ but also ignores self-comparisons among groups under $\text{pairs}(\cdot,\cdot)$ or $\text{compl}(\cdot,\cdot)$:
$$\text{wmean}_{i,j}f_{ij}=\sum_{f_{ij}:\mathcal{S}_i\neq \mathcal{S}_j}(1-|\prob(x\in S_i)-\prob(x\in S_j)|)f_{ij}$$

\section{The FairBench Library}\label{fairbench}
We implement our bias measure definition framework within an open-source Python library called FairBench.\footnote{The documentation of FairBench is available at: \url{https://fairbench.readthedocs.io}} This focuses on ease-of-use, comprehensibility, and compatibility with popular working environments. To achieve these, we adopted a forward-oriented paradigm \cite{krasanakis4180025forward} to design the programming interfaces of code blocks as callable methods that can be combined in reporting mechanisms. Block parameters are transferred through keyword arguments.

\subsection{Forks of sensitive attribute values} 
Our design is centered around uniformly representing many groups of data samples. Previous fairness libraries tend to parse lists of sensitive attribute names and match these with columns of clearly understood programming datatypes, such as Pandas dataframes \cite{mckinney2011pandas}. However, coupling data loading and fairness assessment creates inflexible usage patterns and source code that is harder to extend. For example, it needs new methods and classes to support graph or image AI.

To bypass this issue, FairBench parses vectors of predictions (e.g., scores), ground truth (e.g., classification labels), and binary membership to protected groups. Vectors can come from any computational backend with a duck-type extension of Python's \mintinline{Python}{Iterable} interface; they could be NumPy arrays \cite{harris2020array}, PyTorch tensors \cite{paszke2019pytorch}, TensorFlow tensors \cite{abadi2016tensorflow}, JAX tensors \cite{jax2018github}, Pandas columns \cite{mckinney2011pandas}, or Python lists. We support end-to-end integration with automatic differentiation frameworks by extending their common functional interface provided by the EagerPy library \cite{rauber2020eagerpy} and setting the internal computational backend per \mintinline{Python}{fairbench.backnend(name)}.

We simplify handling of multiple protected groups by organizing them into one data structure, which we call a \mintinline{python}{Fork} of the sensitive attribute. This is a programming equivalent of $\mathbb{S}$. All interfaces accept a sensitive attribute fork, and base measures run for every group. Thus, every group being analysed becomes a computational branch of the fork. We provide dynamic constructor patterns to easily declare forks in common setups. One pattern is demonstrated below, where a fork \mintinline{Python}{s} lets us compute the accuracy fork \mintinline{Python}{acc} that holds assessment outcomes for both whites and blacks:

\begin{lstlisting}[language=Python]
import fairbench as fb
race, predictions, labels = ...  # arrays, tensors, etc
s = fb.Fork(white=..., black=...) # branches hold binary iterables
acc = fb.accuracy(predictions=..., labels=..., sensitive=s)
fb.visualize(acc)  # visualize accuracy, which is also a fork
\end{lstlisting}

Forks may also be constructed from categorical iterables (e.g.,  Pandas dataframe columns or native lists like \mintinline{Python}{cats=['man', 'woman', 'man', 'nonbinary']}) with the constructor pattern \mintinline{Python}{fb.Fork(fb.categories@cats)}. Forks of multidimensional sensitive attributes can be created by adding all categorical parsing to the constructor as positional arguments. Intersections between fork branches per Equation~\ref{cintersect} can be achieved by calling a corresponding method:
\begin{lstlisting}[language=Python]
races, genders = ...  # categorical iterables
s = fb.Fork(fb.categories@races, fb.categories@genders)
s = s.intersectional()
print(s.sum())  # visualization is not very instructive for many branches
\end{lstlisting}

\subsection{Fairness reports}
Multidimensional fairness reports for several default popular measures, comparisons, and reductions can be computed and displayed with the following snippet:
\begin{lstlisting}[language=Python]
vsany = fb.unireport(predictions=..., labels=..., sensitive=s)
pairs = fb.multireport(predictions=..., labels=..., sensitive=s)
report = fb.combine(pairs, vsany)
fb.describe(report)  # or print or fb.visualize
\end{lstlisting}
This combines the comparisons of Equation~\ref{vsAny} (unireport) and Equation~\ref{allvsall} (multireport). For more report types or how to declare one specific measure of bias, refer to the library's documentation. The base measures to analyse are determined from keyword arguments. For example, the above code snippet computes classification base measures, but if we added a \mintinline{Python}{scores=...} arguments (in addition to or instead of \mintinline{Python}{predictions}) recommendation measures would be obtained too. Reports can also parse a custom selection of base measures, including externally defined ones. 

\begin{figure}[htpb]
    \centering
    \includegraphics[width=\textwidth]{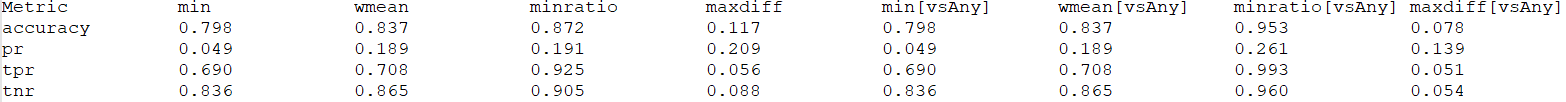}
    \caption{Example of a combined FairBench report.}
    \label{fig:combined}
\end{figure}

The outcome of running the above code is presented in Figure~\ref{fig:combined}. Column names correspond to the combination of reduction and comparison mechanisms, whereas rows to the base measures. Familiarization with both the library and our mathematical framework is required to understand FairBench reports, but these present -to our knowledge- the first systematic way of looking at  many bias assessments at once, regardless of the exact setting. For example, in the report below it is easy to detect the small minimum ratio (minratio) of pr, which corresponds to $\text{prule}$, even if there are generally balanced tpr and tnr (equivalent to balanced fnr and fpr).

\begin{figure}[htpb]
    \centering
    \includegraphics[width=0.54\textwidth]{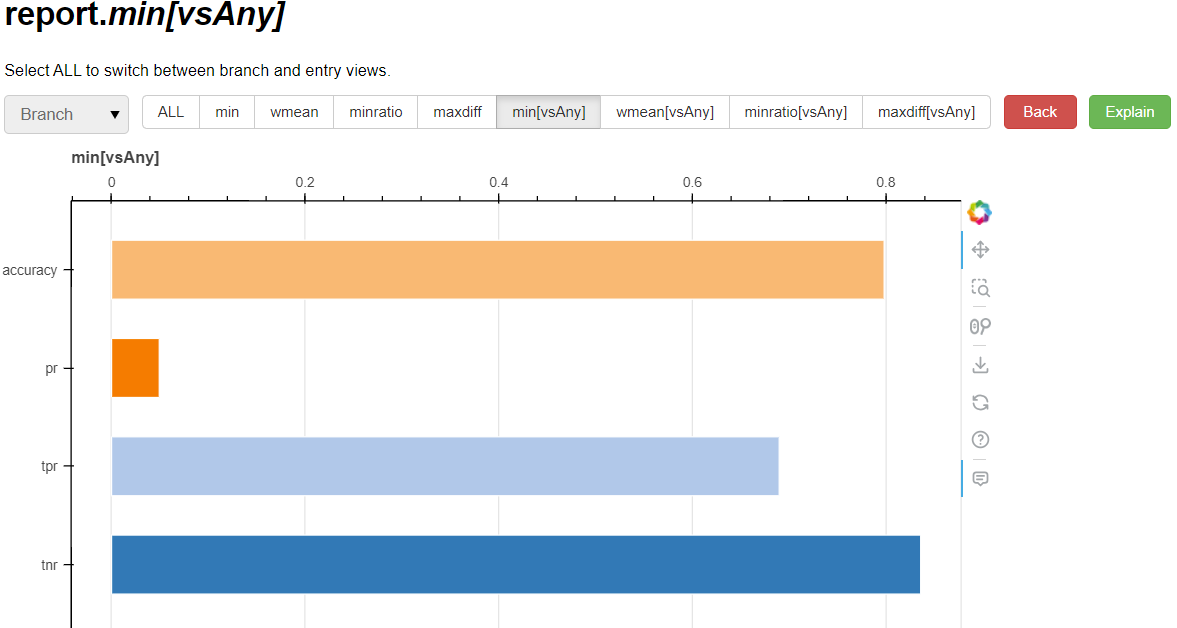}
    \includegraphics[width=0.42\textwidth]{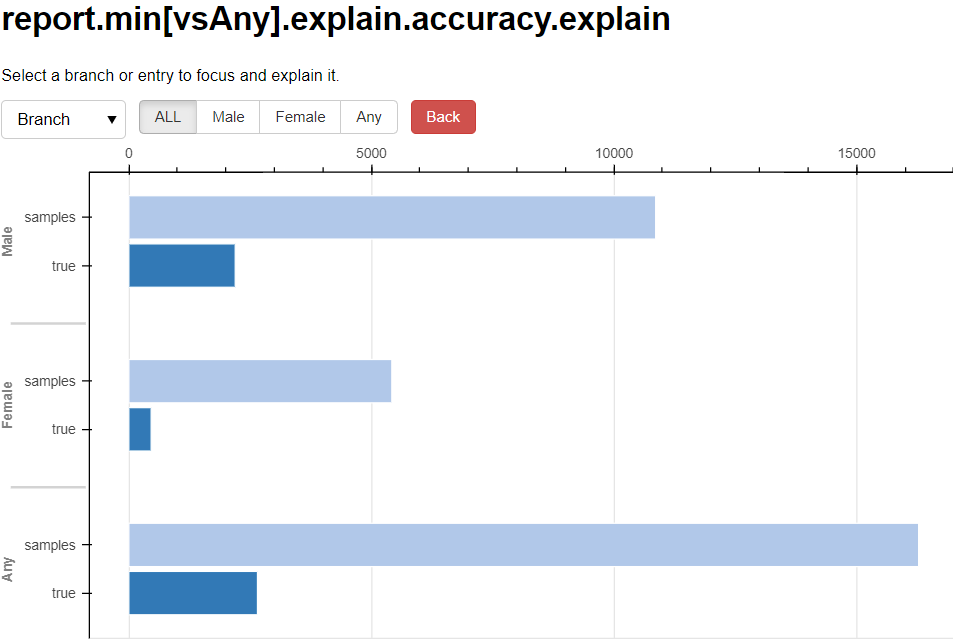}
    \caption{Example steps during the interactive exploration of a FairBench report.}
    \label{fig:interactive}
\end{figure}

Finally, reports can be interactively explored through \mintinline{Python}{fb.interactive(report)}. This uses the Bokeh library \cite{bokeh2014bokeh} to run in notebook outputs or new browser tabs and create visualization similar to Figure~\ref{fig:interactive}. Through the visual interface, users can not only see report values but also focus on columns or rows, and delve into explanations of which raw values contributed to computations, as indicated via the \mintinline{Python}{.explain} part of the exploration path on top. Explanations correspond to tracking metadata values through predicates per Subsection~\ref{base measures}. The same exploration can be performed purely programmatically too.

\section{Conclusions}\label{conclusions}
In this work we provided a mathematical framework that standardizes the definition of many existing and new measures of bias by splitting them in base building blocks and compiling all possible combinations. This framework systematically explores AI fairness through measures that account for a wide range of settings and bias concerns, beyond the confines of ad-hoc exploration. Implementation of popular building blocks are provided in the FairBench library via interoperable Python interfaces. The library provides additional features, like visualization and computation backtracking, that enable in-depth exploration of a wide range of bias concerns. It can also be used alongside popular computational backends.

Future research and development could work towards identifying more bias building blocks by decomposing more measures from the literature, as well as additional block types that may arise in the future. FairBench is open source and we also encourage implementations of such analysis from the community. We further plan to extend currently experimental features, like creating fairness model cards that include caveats and recommendations obtained from interdisciplinary collaborations with social scientists, and creating higher-level assessments.

\section*{Acknowledgement}
This research work was funded by the European Union under the Horizon Europe MAMMOth project, Grant Agreement ID: 101070285. UK participant in Horizon Europe Project MAMMOth is supported by UKRI grant number 10041914 (Trilateral Research LTD).

\bibliography{main}

\end{document}